\documentclass[11pt]{article}

\usepackage[utf8]{inputenc}
\usepackage[T1]{fontenc}
\usepackage[margin=1in]{geometry}
\usepackage{hyperref}
\usepackage{amsmath,amssymb}
\usepackage{graphicx}
\usepackage{booktabs}
\usepackage[expansion=false]{microtype}
\usepackage[numbers]{natbib}
\usepackage{placeins}

\hypersetup{
    colorlinks=true,
    linkcolor=black,
    citecolor=blue,
    urlcolor=blue,
}

\title{\textbf{Code Correctness Is Linearly Decodable from LLM Hidden States Before Generation}}
\author{Carlo Di Cicco\thanks{Independent researcher. Contact: \texttt{cdicicco2001@gmail.com}. Code, data, and analysis scripts: \url{https://github.com/CarloDiCicco/ReasoningLab}}}
\date{}

\begin{document}

\maketitle

\begin{abstract}
Large language models encode rich information in their hidden states. 
This work asks whether the correctness of code that Qwen3-4B-Instruct-2507 has not yet generated is already legible in its hidden states, evaluated on a set of 444 tasks from LiveCodeBench.
The correctness of the model's first-attempt code is linearly decodable from the hidden state at the final prompt token, captured before any output token is generated, with a leakage-free held-out AUC of $0.881 \pm 0.008$ across 50 outer splits. 
To assess whether this signal is explained by prompt length, each hidden state dimension is residualized with respect to its linear effect. The probe still achieves an AUC of $0.842 \pm 0.010$, substantially above a logistic prompt-length baseline of $0.657 \pm 0.014$, and none of the nonlinear models tested improves upon it. 
A companion question about whether self-repair leaves a geometric signature in the model's hidden states could not be answered, because successful repairs following a failed first attempt are too rare in this setting to support the analysis. 
The contribution is both empirical and methodological, providing evidence that pre-generation hidden states contain a robust signal of eventual code correctness, together with a confound-control diagnostic that quantifies how much of that signal survives adjustment for prompt length.
\end{abstract}

\section{Introduction}
\label{sec:introduction}

Reasoning has become a central axis of large language model development. With each generation, models train on more and better data with improved training methods, and become capable of tackling problems that would have been out of reach a few years ago. Their internals, in turn, grow more informative: a richer internal computation is doing the work behind richer external behavior, which makes the question of what those internals encode increasingly worth asking. Code generation is a particularly clean field for that question: outputs are easily verifiable against unit tests, with no human-judgment loop required to decide whether an answer is correct. As model performance on code-generation benchmarks has improved rapidly, the question is no longer only \emph{whether} a model can produce correct code, but also \emph{what happens internally} as it does so. This shifts the focus toward mechanistic interpretability in models that now exhibit genuine capability in code generation.

A standard tool for asking what is encoded inside a frozen language model is the \emph{linear probe}: a simple linear classifier trained on a hidden state vector to predict some property of interest, where probe performance is read as a measure of how linearly decodable that property is from the representation \citep{alain2016probes}. Two recent works have studied code correctness through hidden states \citep{bui2025openia, ribeiro2025internal}, reading hidden states of already-generated code to judge whether it is correct without running the unit tests. One trains a classifier on those states; the other contrasts correct and incorrect samples to extract a correctness direction and rank candidate generations. In both, the hidden state is read \emph{after} the model has produced an answer.

This work asks a related but earlier question: whether correctness is already decodable from the hidden state at the position of the last prompt token, captured on a single forward pass over the prompt and before any output token is sampled. Whereas those works operate on hidden states of generated code, the probe in this paper operates on the hidden state of the prompt alone. This is a more upstream interpretability claim: the information must already be present at the moment the model finishes reading the input, not just at the moment it has committed to a particular output.

The finding is the following. The pre-generation probe holds up cleanly: the prompt-final hidden state carries a linearly decodable correctness signal that is robust across many random train/test splits and that survives controlling for prompt length.
A companion question, whether the hidden state shift from a failing attempt to its repair attempt carries a geometric signature of repair success, could not be answered on this data.
Prior reports find that iterative self-repair plateaus after the first one or two attempts \citep{arimbur2026howmanytries}, and the collapse here is even sharper: the pass rate drops steeply as soon as the first attempt fails, and successful repairs are too rare at every transition to support a direction analysis. Section~\ref{sec:limitations} reports this as a null result.
A probe trained on raw hidden states could in principle read off superficial properties of the input rather than anything mechanistic about the model's representation, and prompt length turns out to be exactly such a property: used as the only feature, it already predicts whether the first attempt will pass better than random guessing.
Residualization, the operation of regressing each hidden state dimension on the candidate covariate and subtracting the prediction, removes the part of the signal the covariate can account for, and what remains after the removal is what the probe can genuinely claim.
Residualizing against prompt length confirms that the decodable correctness signal is not reducible to this factor, even if a small part of the raw probe's performance is. The procedure measures how much of a raw signal is genuinely its own.
\section{Setup}
\label{sec:setup}

\subsection{Model and data}

All experiments use \texttt{Qwen/Qwen3-4B-Instruct-2507} \citep{qwen3report2025}, a 36-layer decoder-only transformer with a 2560-dimensional hidden state. The dataset consists of 444 LiveCodeBench \citep{jain2024livecodebench} tasks spanning the easy, medium, and hard difficulty tiers. Every experiment runs on a single NVIDIA DGX system.

\subsection{Hidden state capture}

Across all task attempts, the hidden state at the final prompt token is extracted from a prompt-only forward pass, before the autoregressive continuation begins. This yields a single 2560-dimensional vector per task per attempt at each captured layer. Hidden states are captured for the upper transformer blocks (layers 29--36), motivated by the general finding that linear separability of high-level properties increases with depth in deep networks \citep{alain2016probes}. Layer selection is handled by nested cross-validation (CV) in Section~\ref{sec:probe_results}.

\subsection{Repair-B policy}

For tasks where the model's first attempt fails the unit tests, an iterative repair procedure (``repair-B'') is invoked, with a maximum of 5 total attempts per task. Each attempt is a fresh forward pass on the full prompt, with no state carried over from previous attempts, and the repair prompt contains the original problem, the model's previous code, the error output captured by the verifier, and an instruction to fix the failure. Generation is capped at 1024 new tokens per attempt. The pre-generation probe in Section~\ref{sec:probe_results} uses all 444 tasks at attempt 0, and the outcome of the repair procedure itself is summarized in Section~\ref{sec:limitations}.

\section{Correctness is linearly decodable before generation}
\label{sec:probe_results}
Before any token is generated, the hidden state at the last prompt token encodes information about whether the model's first-attempt generation will be correct. A linear probe trained on this single vector predicts pass/fail on held-out tasks with mean test area under the ROC curve (AUC) of $0.881 \pm 0.008$ over $50$ random splits, well above a prompt-length baseline of $0.657 \pm 0.014$. The full procedure is explained below.

\subsection{Procedure}

For each of the 444 LiveCodeBench tasks, the hidden state at the last prompt token of the first attempt (attempt 0) is captured, producing a single 2560-dimensional vector per task per layer, as described in Section~\ref{sec:setup}. Each vector is labeled by whether the attempt-0 code passed the task's unit tests, and the probe is trained to predict this label. The pipeline is a three-step composition:
\begin{enumerate}
    \item Standardization (mean-centering and unit-variance scaling), fit on the training fold only.
    \item PCA retaining 95\% of cumulative variance, fit on the training fold only.
    \item Logistic regression with $\ell_2$ regularization. The regularization strength $C$ is selected by 5-fold CV inside the training fold, scoring on AUC, over the grid
    \[
        \{10^{-3}, 10^{-2}, 10^{-1}, 1, 10\}.
    \]
\end{enumerate}

All three pipeline components are implemented in scikit-learn \citep{pedregosa2011sklearn}.

To get a stable estimate rather than rely on a single split, the pipeline is fit and evaluated $50$ times under different random stratified $80/20$ splits, and performance is summarized by the mean test AUC with a $95\%$ confidence interval. Three configurations are compared: two apply the three-step pipeline above to different feature sets, and the third is a simpler prompt-length baseline.
\begin{enumerate}
    \item \textbf{Raw}: the three-step pipeline applied directly to the hidden state vector as captured.
    \item \textbf{Residualized}: for each of the 2560 hidden state dimensions, an OLS regression is fit on the train fold with that dimension as response and \texttt{prompt\_tokens}, the number of tokens in the prompt, as the single predictor. The fitted prediction is subtracted from both train and test for that dimension, and the standard three-step pipeline is applied to the residualized features. Residualization is therefore re-fit on every random split.
    \item \textbf{Prompt-length baseline}: a logistic regression on \texttt{prompt\_tokens} alone (a single feature, so no dimensionality reduction such as PCA is needed), fit and evaluated under the same 50 splits with $C$ selected by cross-validation as above.
\end{enumerate}

The residualized configuration and the prompt-length baseline both target the same confound. Prompt length alone predicts the attempt-0 label above chance, as the baseline below shows, so a probe trained on raw hidden states could in principle exploit it by reading off prompt length rather than anything mechanistic about the model's internal representation. Residualizing each hidden state dimension against \texttt{prompt\_tokens} removes the linear contribution of prompt length from every dimension, leaving only the part of the hidden state that is not linearly explained by it. The prompt-length baseline goes one step further: it asks how well a probe can do using \texttt{prompt\_tokens} \emph{alone}, giving a reference the residualized probe must clear to show it carries signal beyond length.

The single sample per task ensures no task appears in both train and test splits, removing the inflation that would arise from leakage across attempts of the same problem.

\subsection{Layer landscape and selection}
\label{subsec:layer_selection}

The raw three-step pipeline is applied to each of the captured upper-block layers (29--36). Running it independently per layer, on 50 random stratified $80/20$ splits each, gives the descriptive landscape in Table~\ref{tab:layer_sweep}, a gentle plateau that peaks near the shallow end of the captured range at layer 30 and declines slightly toward layer 36. Unlike the monotonic increase of linear separability with depth reported for convolutional networks \citep{alain2016probes}, the probe AUC here peaks before the final block and then declines, suggesting that final transformer layers shift from maintaining general semantic representations to optimizing for next-token prediction.

\begin{table}[h]
\centering
\begin{tabular}{lcccccccc}
\toprule
Layer & 29 & 30 & 31 & 32 & 33 & 34 & 35 & 36 \\
\midrule
Test AUC & $0.883$ & $0.884$ & $0.882$ & $0.882$ & $0.883$ & $0.882$ & $0.880$ & $0.880$ \\
CV AUC   & $0.877$ & $0.878$ & $0.876$ & $0.876$ & $0.877$ & $0.875$ & $0.873$ & $0.873$ \\
\bottomrule
\end{tabular}
\caption{Per-layer probe performance on the 444 LiveCodeBench tasks: an independent per-layer sweep, using its own 50 random stratified 80/20 splits, each layer is evaluated on both mean test and CV AUC. The upper blocks form a shallow plateau peaking at layer 30. This table is descriptive only, selecting the argmax layer on these test scores would constitute selection on the test set, which the nested CV procedure below avoids.}
\label{tab:layer_sweep}
\end{table}

Because the layer is a hyperparameter, choosing it by the test-set argmax of Table~\ref{tab:layer_sweep} would leak the test set into the model-selection step and bias the reported performance upward \citep{cawley2010overfitting}. The layer is therefore selected by nested CV, following the same validation-based layer-selection logic used in prior work on internal correctness representations \citep{ribeiro2025internal}. Across 50 outer stratified $80/20$ splits, the layer, jointly with the regularization strength $C$, is selected by 5-fold CV within each outer training fold; the selected configuration is refit on the full outer-training fold and evaluated once on the untouched outer-test fold. Selection spreads across the captured range, with layer 30 the modal choice at $16$ of $50$ outer splits, layer 29 next at $13$, and layer 33 at $8$, as Table~\ref{tab:nested_cv_selection} shows. The modal selection across the 50 splits is layer 30 with $C = 0.001$, chosen in $16$ of them. The leakage-free outer-test AUC of the full select-then-train procedure is $0.881 \pm 0.008$. The descriptive sweep in Table~\ref{tab:layer_sweep} is consistent with this selection, peaking at layer 30 with the neighboring layers statistically indistinguishable.

\begin{table}[h]
\centering
\begin{tabular}{lcccccccc}
\toprule
Layer & 29 & 30 & 31 & 32 & 33 & 34 & 35 & 36 \\
\midrule
Times selected (of 50) & $13$ & $16$ & $4$ & $3$ & $8$ & $2$ & $2$ & $2$ \\
\bottomrule
\end{tabular}
\caption{Nested-CV layer selection, how often each layer is chosen, across all $C$ values, as part of the best (layer, $C$) configuration over the 50 outer splits. The grid search operates jointly over layer and $C$, and this table shows the layer counts collapsed across $C$ to make the layer preference visible. Selection spreads across the captured range, with layer 30 the mode at $16$ splits and layer 29 next at $13$.}
\label{tab:nested_cv_selection}
\end{table}

\FloatBarrier
\subsection{Results}

Each number in Table~\ref{tab:probe_results} is a leakage-free procedure-level estimate: for the raw and residualized probes, layer and $C$ are jointly selected inside each outer training fold and evaluated on the untouched outer-test fold; for the prompt-length baseline, only $C$ is selected as there is no layer. Reporting procedure-level numbers for all three configurations ensures they are directly comparable, since each reflects the performance of the full select-then-fit pipeline applied consistently, rather than locking any configuration to a specific hyperparameter value chosen by inspecting the test set.

\begin{table}[h]
\centering
\begin{tabular}{lc}
\toprule
Configuration & Test AUC (mean $\pm$ 95\% CI) \\
\midrule
Raw hidden states    & $0.881 \pm 0.008$ \\
Residualized         & $0.842 \pm 0.010$ \\
Prompt-length baseline & $0.657 \pm 0.014$ \\
\bottomrule
\end{tabular}
\caption{Nested CV procedure-level performance for the three probe configurations, averaged over 50 outer stratified $80/20$ splits. Values are leakage-free outer-test AUCs.}
\label{tab:probe_results}
\end{table}

The raw configuration reaches mean outer-test AUC $0.881$. After linearly removing prompt length from every hidden state dimension, the residualized configuration reaches $0.842$, a drop of only $0.039$ from raw. The prompt-length baseline reaches $0.657$. The gap between the residualized probe and the baseline is about $19$ AUC points, placing a concrete floor under the result, a substantial portion of the hidden state's pass/fail signal is not explained by prompt length and survives its removal. The signal is present before a single output token is generated, encoded in the model's internal representation of the prompt alone.

The prompt-length baseline reported above uses logistic regression. To check that this floor does not depend on the choice of a linear model, a random forest, a gradient boosting model, and a small multilayer perceptron were each fit on \texttt{prompt\_tokens} alone under the same nested CV procedure described in Section~\ref{subsec:layer_selection}. Table~\ref{tab:length_baselines} reports their mean outer-test AUC. None exceeds the logistic baseline. Even fit with nonlinear models, prompt length does not approach the residualized probe, so the probe's gap is not an artifact of the baseline being restricted to a linear model.

\begin{table}[h]
\centering
\begin{tabular}{lc}
\toprule
Prompt-length model & Test AUC (mean $\pm$ 95\% CI) \\
\midrule
Logistic regression   & $0.657 \pm 0.014$ \\
Random forest         & $0.642 \pm 0.013$ \\
Gradient boosting     & $0.649 \pm 0.015$ \\
Multilayer perceptron & $0.657 \pm 0.014$ \\
\bottomrule
\end{tabular}
\caption{Mean outer-test AUC for prompt-length models, each fit on \texttt{prompt\_tokens} alone following the nested CV procedure of Section~\ref{subsec:layer_selection}, with model-specific hyperparameter grids tuned in the inner folds. The nonlinear families do not improve on the logistic model.}
\label{tab:length_baselines}
\end{table}

\FloatBarrier

\section{Limitation and null result}
\label{sec:limitations}

This section reports one limitation and one null result of the study.

\begin{itemize}
    \item \textbf{Single model.} All experiments use Qwen3-4B-Instruct-2507. Replicating across model families would require substantially more inference compute and time, this is left for future work.

    \item \textbf{Repair geometry is a null result by data starvation.} A companion question asked whether the hidden state shift from a failing attempt to its repair attempt carries a geometric signature of repair success. The data cannot support it. Of the 266 tasks whose first attempt fails, only 20 ever recover within the 5-attempt budget, 14 of them at the first repair, and the conditional pass rate among still-unsolved tasks collapses from 40.1\% at attempt 0 to 5.3\%, 2.0\%, 0.4\% and 0\% at attempts 1 through 4. Prior reports find that iterative self-repair plateaus after the first one or two attempts \citep{arimbur2026howmanytries}, and the decay here is even sharper, with the pass rate collapsing at the very first repair. No contrastive direction can be meaningfully estimated from 14 successful repairs in a 2560-dimensional space, so no claim about repair geometry is made in either direction, and the distinction between ``the signal does not exist'' and ``the data were not sufficient to detect it'' cannot be resolved with the present dataset.
\end{itemize}

\section{Conclusion}
\label{sec:conclusion}

This work examined whether code correctness is decodable from the prompt-final hidden states of Qwen3-4B-Instruct-2507 on LiveCodeBench tasks, before any token is generated.
A linear probe answered yes, with performance robust across many random splits and surviving the linear removal of prompt length: the residualized probe reaches $0.842$ AUC against a prompt-length baseline of $0.657$.
Prompt length alone, fit with nonlinear models as well as a linear one, stayed far below the probe, so its lead is not an artifact of a linear baseline.
A companion question, whether the hidden state shift from a failing attempt to its repair carries a geometric signature of repair success, could not be answered: successful repairs are too rare in this setting to estimate any direction, and Section~\ref{sec:limitations} reports it as a null result.

The probe's signal survived residualizing against prompt length while giving up only a small part of its raw performance, showing that the hidden states carry a correctness signal not reducible to input length.
The value of residualization is that it separates the part of a raw signal the covariate can account for from the part that is genuinely the representation's own, and reports only the latter.

Two extensions of this work stand out. The first is to replicate the analysis across other open-weight models, which would help separate findings that are specific to this one model from properties shared across language models.
The second is to gather many more successful self-repairs, for instance on easier tasks or with a stronger model, on which the repair-direction question left open here could actually be tested.
Code generation is a particularly tractable domain for this kind of analysis: outputs are objectively verifiable, the failure modes are observable, and each task gives a binary pass/fail signal. More generally, as language models grow more capable their internal computation grows more elaborate, and there is correspondingly more to learn from studying what those internals encode.
The contribution of this study is as much methodological as empirical. A signal is only as trustworthy as the confound control it survives, and what this paper claims is exactly what survived it.

\bibliographystyle{unsrtnat}
\bibliography{references}

\end{document}